\documentclass{article}

\usepackage{microtype}
\usepackage{graphicx}
\usepackage{subcaption}
\usepackage{booktabs}
\usepackage{tabularx}
\usepackage{hyperref}
\usepackage{amsmath}
\usepackage{amssymb}
\usepackage{mathtools}
\usepackage{amsthm}
\usepackage{enumitem}
\usepackage[capitalize,noabbrev]{cleveref}
\PassOptionsToPackage{numbers,sort&compress,square}{natbib}
\usepackage[preprint]{icml2026}
\bibpunct{[}{]}{,}{n}{}{,}

\makeatletter
\renewcommand{\printAffiliationsAndNotice}[1]{%
  \global\icml@noticeprintedtrue
  \stepcounter{@affiliationcounter}%
  {\let\thefootnote\relax
   \footnotetext{%
     \noindent\raggedright
     \ificmlshowauthors #1\par\fi
     \forloop{@affilnum}{1}{\value{@affilnum} < \value{@affiliationcounter}}{%
       \textsuperscript{\arabic{@affilnum}}%
       \ifcsname @affilname\the@affilnum\endcsname
         \csname @affilname\the@affilnum\endcsname
       \else
         {\bf AUTHORERR: Missing \textbackslash icmlaffiliation.}%
       \fi
       \par
     }%
     \ifdefined\icmlcorrespondingauthor@text
       Correspondence to: \icmlcorrespondingauthor@text.\par
     \fi
     \Notice@String
   }%
  }%
}
\makeatother
\icmltitlerunning{Respiratory Status Detection with Video Transformers}

\begin{document}

\twocolumn[
  \icmltitle{Respiratory Status Detection with Video Transformers}

  \begin{icmlauthorlist}
    \icmlauthor{Thomas Savage}{aff1}
    \icmlauthor{Evan Madill}{aff2}
  \end{icmlauthorlist}

  \icmlaffiliation{aff1}{University of Pennsylvania}
  \icmlaffiliation{aff2}{University of Washington}
  \icmlcorrespondingauthor{Thomas Savage}{thomassavage131@gmail.com}

  \icmlkeywords{video transformers, respiratory distress, computer vision, ViViT, LieRE, motion-guided masking}

  \vskip 0.3in
]

\printAffiliationsAndNotice{}

\begin{abstract}
Recognition of respiratory distress through visual inspection is a life-saving clinical skill.  Clinicians can detect early signs of respiratory deterioration, creating a valuable window for earlier intervention.   In this study, we evaluate whether recent advances in video transformers can enable Artificial Intelligence systems to recognize the signs of respiratory distress from video.  We collected videos of healthy volunteers recovering after strenuous exercise and used the natural recovery of each participant’s respiratory status to create a labeled dataset for respiratory distress.  Splitting the video into short clips, with earlier clips corresponding to more shortness of breath, we designed a temporal-ordering challenge to assess whether an AI system can detect respiratory distress.  We found a ViViT encoder augmented with Lie Relative Encodings (LieRE) and Motion-Guided Masking, combined with an embedding-based comparison strategy, can achieve an F1 score of 0.81 on this task.  Our findings suggest that modern video transformers can recognize subtle changes in respiratory mechanics.  
\end{abstract}

\section{Introduction}

Respiratory status evaluation is a key skill taught to all nurses, physicians, and paramedics. Signs of distress such as nasal flaring, skin retraction, and the use of accessory muscles can be subtle, but indicate that a patient is at high risk for respiratory failure and even death \cite{shah2017child,hedstrom2018silverman,maitre1995physical}. Providers become attuned to these signs over years of training, which enables them to look beyond telemetry or laboratory-based measurements of respiratory status and recognize impending decompensation.  In turn, frequent monitoring of high-risk patients by trained professionals is both a significant safety priority and operational expense for hospitals and skilled nursing facilities\cite{cfr42publichealth,cfr48335,bls2023snf,harrington2024nursing}.

In this paper, we investigate whether a computer vision application can be trained to recognize the subtle signs of respiratory distress.  Historically, computer vision applications have struggled with this task because the detection of small changes in breathing mechanics is confounded by patient motion, body positioning, and variable lighting conditions \cite{akamatsu2024calibrationphys,chen2024actnet}. Encouragingly, recent advances in video transformer positional encoders have improved spatiotemporal modeling and may enable detection of these subtle respiratory changes \cite{ostmeier2025liere}.

Building on these advances, our objective is to determine whether modern video transformer architectures can detect subtle changes in breathing mechanics. We evaluate this hypothesis in a cohort of healthy volunteers who performed exercise and recorded their respiratory recovery.  We trained the model on a pairwise temporal ordering task, predicting which of two short clips from the same recovery sequence occurred earlier (greater respiratory distress) versus later (closer to baseline).  Accurate ordering of these clips provides evidence that computer vision models can capture subtle changes in respiratory status.

\section{Related Work}

\subsection{Video Transformer Background}

Video transformer architectures such as ViViT \cite{arnab2021vivit} and TimeSformer \cite{bertasius2021timesformer} represent videos as sequences of spatiotemporal patch tokens, using attention to model relationships across space and time for video classification. Because attention alone has no built-in awareness of token position, these models must provide positional information to distinguish the location of tokens. Early video transformers used absolute positional embeddings (APEs), which provide each token a learned position label.  A limitation of APEs is that they can lead to position-specific overfitting, thereby degrading performance, especially for tasks that rely on subtle, temporally-evolving visual signals \cite{sinha2021absolute}.

To improve performance, relative positional encodings were developed to reduce the risk of overfitting. Rotary Position Embeddings (RoPE) is a widely used relative method that encodes position by applying position-dependent rotations to the query and key vectors, so their dot-products naturally reflect relative token position rather than requiring learned absolute position tokens \cite{su2023roformer}. Although RoPE was originally developed for 1D sequences, such as language models, rotary approaches have since been adapted to higher-dimensional vision settings.  Lie Relative Encodings (LieRE) further generalizes rotary encodings by learning higher-dimensional rotation bases for multi-axis coordinate systems (e.g., x, y, and time for video data), enabling richer spatiotemporal position representations and significant improvements in performance \cite{ostmeier2025liere}.

An additional strategy that informed our approach is Motion-Guided Masking (MGM), which encourages the model to focus on regions of the video that exhibit meaningful motion \cite{huang2023mgmae}. In MGM, each frame is divided into a grid of patches, optical flow based motion is computed for each patch, and only the most dynamic patches are retained as input to the model. The remaining patches are masked (blacked out). By restricting focus to the most informative spatiotemporal regions, this motion-guided masking helps the transformer down-weight irrelevant visual content (e.g., background, lighting variation, or non-respiratory movement) and can improve performance on tasks that depend on subtle temporal changes \cite{choudhury2024dontlooktwice,huang2023mgmae}.

\subsection{Respiratory Distress Assessment via Computer Vision}

Most published computer vision methods for respiratory monitoring focus on estimating respiratory rate or tidal volume, rather than detecting the classic visual physical exam signs of respiratory distress \cite{akamatsu2024calibrationphys,chen2024actnet,bai2010embedded}. In our literature review, we identified only one study—Nawaz et al. \cite{nawaz2024automated}---that explicitly targeted physical exam signs of dyspnea. In their work, the authors used a 3D convolutional neural network to identify chest retractions and detect acute respiratory distress in children admitted to a pediatric intensive care unit.

A key limitation of the Nawaz et al. approach was its reliance on manual video cropping and clinician-guided positioning prior to recording \cite{nawaz2024automated}. While this added supervision can improve signal quality, it reduces scalability in real-world monitoring environments, where deployment typically requires minimal curation and limited dependence on human-in-the-loop preprocessing.

In contrast, our experiment focuses on fully automated video processing without manual cropping or clinician-guided framing. This design choice prioritizes practicality for routine monitoring settings and is better suited for scalable deployment.

\section{Methods}

In this study, we evaluate respiratory status change detection through a pairwise temporal ordering objective. Given two short video clips sampled from the same participant’s post-exercise recovery recording, the model predicts which clip occurred earlier in the sequence—corresponding to greater shortness of breath—versus later, when the participant is closer to their baseline. To solve this challenge, we evaluated multiple strategies for (1) comparing clip states and (2) capturing spatiotemporal features for best performance.

\subsection{Clip-State Comparison}

Because our experiment compares respiratory states between two clips, rather than performing standalone classification on a single clip, we required an explicit mechanism to compare two videos. We evaluated two primary approaches:
\begin{itemize}[noitemsep, topsep=0pt]
    \item Two-tower analysis with cross-attention
    \item Mapping each video to an embedding state for comparison
\end{itemize}

The two-tower cross-attention approach processed the two clips through separate model towers, producing two feature representations. These representations were then contrasted using cross-attention, enabling direct interaction between clip representations. We tested two cross-attention configurations: (1) full cross-attention across all layers and (2) cross attention at the level of the CLS-token. The model was trained to predict which of the two clips depicted greater shortness of breath (i.e., whether the first or second clip occurred earlier in the recovery sequence).

In our embedding-state approach, a single model encoded each clip into a fixed-length embedding vector. Because ground-truth shortness of breath labels were not available at the clip level, we used weak supervision to train each clip’s position within the recovery spectrum: clips from the first third of the sequence were assigned a proxy shortness of breath (SOB) label, while clips from the last third were assigned to a no shortness of breath (no SOB) category, and clips from the middle third were excluded to avoid ambiguous intermediate states. During training, we jointly learned two class prototypes (SOB and no SOB) together with the encoder. This objective minimized cosine-distance loss between each clip embedding and its corresponding class prototype. At inference, clips were scored by their distance to the two prototype vectors; embeddings closer to the SOB prototype were interpreted as reflecting a more SOB-like respiratory state.

\subsection{ViViT-Base Augmentation}

Our baseline video encoder was a factorized ViViT-base model. To assess whether improved spatiotemporal representations would improve performance, we evaluated two modifications to the baseline architecture.

First, we replaced the standard absolute positional embeddings with LieRE. LieRE provides a richer positional representation by improving how the model encodes token location and supports reasoning about relative position across the video’s spatial and temporal axes.

Second, we evaluated MGM as an input pre-processing step. Using optical flow, we computed motion magnitude for each patch and retained only the top 20\% most dynamic patches in each frame. All remaining patches were masked to black. This masking strategy was intended to bias the encoder toward regions exhibiting meaningful temporal change and to down-weight largely static content (e.g., background).

\subsection{Model and Training Settings}

Our baseline encoder was a factorized ViViT-style transformer with an embedding dimension of 768, 12 attention heads, 6 temporal transformer layers, and 6 spatial transformer layers. Input clips were standardized to 6 seconds with 32 frames at 224×224 resolution and tokenized using 16×16 patches, resulting in a 14×14 patch grid per frame.

For MGM, we computed optical flow between consecutive frames and aggregated motion magnitude over 16×16 pixel tiles. Masking was then applied at this tile resolution, retaining only the top 20\% most dynamic tiles of each frame.

All models were trained on a single NVIDIA A100 GPU. To keep experiments computationally comparable, training for each model was capped at \textless 100 hours and a maximum of 5 epochs.

\subsection{Participant Cohort}

Study participants were recruited via an online advertisement. Inclusion criteria required that participants be 18 years old, engage in strenuous exercise at least once per week, and self-classify their physical activity level as intermediate or advanced. Key exclusion criteria included any history of cardiac, pulmonary, or other medical conditions that would preclude participation in strenuous exercise.

Eligible participants were instructed to complete 5 minutes of strenuous exercise and then record a 5-minute video of their respiratory recovery. Each participant submitted three videos, recorded using either a laptop or a mobile phone.

In total, 75 participants submitted video recordings. Fifty-two participants were included in the final analysis. The remaining 23 were excluded because either (1) the video resolution was insufficient, (2) the participant was not visible in the frame, or (3) author TS could not reliably determine the presence of shortness of breath after comparing the first 10 seconds of the video to the last 10 seconds.

\subsection{Dataset Construction and Splitting}

Exercise participant data were partitioned into two sets: a training/evaluation set and a held-out test set. The test set consisted of seven videos from three participants. These participants (and all of their videos) were excluded from model development and used only for final testing. The remaining 49 participants comprised the training/evaluation dataset.

Each video was segmented into 6-second clips and cropped automatically, with no manual manipulation or human supervision.

\subsection{Ethics Review}

Exercise participant video data was collected under University of Pennsylvania IRB 858520.

\section{Results}

We evaluated our approach in three steps: (1) comparing mechanisms for paired clip-state discrimination, (2) testing whether ViViT augmentations improved model performance, and (3) examining how model performance varied with the magnitude of respiratory-status differences between clips.

First, we assessed how different architectural mechanisms discriminated respiratory status differences between paired clips. Our embedding-distance approach achieved the best performance with an F1 score of 0.75. Performance was lower with two-tower CLS-token cross-attention (F1 of 0.69) and two-tower full cross-attention (F1 of 0.58) methods. Training time was also substantially lower for the embedding-distance method than for the two-tower approaches. Collectively, these results indicate that an embedding-based comparison is the most effective strategy for paired clip comparison in this respiratory status comparing task.  Given this result, we used the embedding-distance approach for all subsequent experiments.  

\begin{table}[b]
    \caption{Accuracy and F1 score comparison for tested clip-state comparison methods as well as ViViT augmentation mechanisms.  
*TT represents a Two-Tower model architecture with cross-attention at the specified level.
}
    \label{tab:main_results}
    \centering
    \small
    \begin{tabularx}{\columnwidth}{>{\raggedright\arraybackslash}X >{\raggedright\arraybackslash}X c c}
        \toprule
        Clip Comparison Mechanism & ViViT Augmentation & Accuracy & F1 \\
        \midrule
        TT - Full & -- & 60.2\% & 0.58 \\
        TT - CLS-Token & -- & 58.4\% & 0.69 \\
        Embedding dist & -- & 60.1\% & 0.75 \\
        Embedding dist & LieRE & 62.3\% & 0.77 \\
         \textbf{Embedding dist} & \textbf{LieRE + MGM} & \textbf{67.5\%} & \textbf{0.81} \\
        \bottomrule
    \end{tabularx}

\end{table}

We next evaluated whether augmentations to ViViT-base could further improve model performance. Replacing absolute positional embeddings with Lie Relative Encodings (LieRE) increased performance to an F1 score of 0.77.  Then adding Motion-Guided Masking (MGM) further improved performance to an F1 score 0.81.  In turn, the best-performing configuration combined both LieRE and MGM with our embedding-distance approach.

\begin{figure}[t]
    \centering
    \includegraphics[width=\columnwidth]{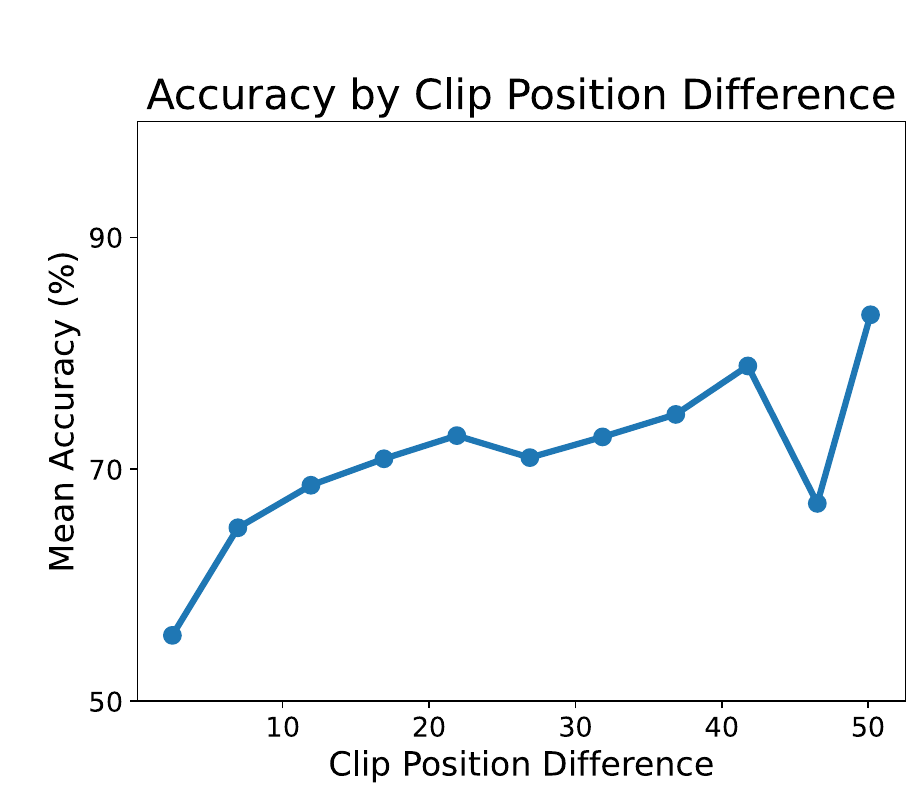}
    \caption{Accuracy analysis for the top performing LieRE-MGM-embedding model as a function of shortness of breath severity.  The larger the difference in clip position, the more likely the model was able to accurately assess the subject’s respiratory status.  A clip position of 1 is equal to 6 seconds difference in a participant's respiratory recovery course.}
    \label{fig:clip_separation}
\end{figure}

Finally, we examined how model performance varied as the degree of respiratory status change increased between clips, proxied by their index separation within the recovery sequence. Accuracy increased as the distance between clips grew; when clips reflected only small differences in respiratory status, accuracy ranged from 55–65\%, whereas comparisons involving larger differences achieved nearly 80\% accuracy  (\Cref{fig:clip_separation}). This pattern was expected, as the visual signs of shortness of breath become progressively less apparent as participants recover from exercise.

\section{Discussion}

The results of our study demonstrate that modern video transformer–based computer vision models can learn to discriminate respiratory states in healthy participants during post-exercise recovery, achieving an F1 score of 0.81 and accuracy of 67.5\%. Across the clip-state comparison strategies we tested, the embedding-distance approach outperformed two-tower cross-attention, while requiring substantially less training time, supporting embedding-based comparison as an effective and efficient framework for paired respiratory-state comparison. Subsequently building on the embedding-distance approach, augmentation with LieRE and MGM yielded the strongest overall results. Taken together, these findings suggest that computer vision models can be designed to capture subtle, visually-expressed differences in respiratory status.

\subsection{Limitations}

Our study has several limitations. Most notably, although we intentionally collected videos without clinician-guided cropping or manual frame selection, the data were still collected under a degree of participant supervision. Participants implicitly curated their recordings by positioning the camera and keeping their face and upper body in view. As a result, our evaluation does not fully represent a passive, continuous monitoring setting.  Future work should therefore evaluate models on datasets collected under truly naturalistic conditions, where patients are recorded in the course of their daily routine or standard care.

Another limitation of our experimental design is the potential confounding of perspiration after exercise.  The model may be able to recognize small differences in perspiration and draw indirect conclusions regarding the patient's shortness of breath state.  Future work with a real world patient dataset will be needed to account for this potential confounder.

\section{Conclusion}

This study demonstrates computer vision models can be trained to detect subtle changes in respiratory status in healthy participants during post-exercise recovery. Future work should focus on building real world patient datasets and evaluating whether modern methods with MGM, LieRe position embeddings and an embedding-state clip comparison mechanism can detect and track pathologic dyspnea under clinical conditions.

\end{document}